\documentclass[10pt,twocolumn,letterpaper]{article}

\usepackage{iccv}
\usepackage{times}
\usepackage{epsfig}
\usepackage{graphicx}
\usepackage{amsmath}
\usepackage{amssymb}

\usepackage[font=small]{caption}
\usepackage{subcaption}
\usepackage{booktabs}
\usepackage{mathtools}
\usepackage{enumitem}
\usepackage{arydshln}

\usepackage[pagebackref=true,breaklinks=true,letterpaper=true,colorlinks,bookmarks=false]{hyperref}

\iccvfinalcopy 

\def\iccvPaperID{649} 

\ificcvfinal\fi
\begin{document}

\title{ScaleNet: Guiding Object Proposal Generation in Supermarkets and Beyond}

\author{Siyuan Qiao$^{1}$ \hskip 1em Wei Shen$^{1,2}$ \hskip 1em Weichao Qiu$^{1}$ \hskip 1em Chenxi Liu$^{1}$ \hskip 1em Alan Yuille$^{1}$\\
Johns Hopkins University$^{1}$ \hskip 1em Shanghai University$^{2}$\\
{\tt\small \{siyuan.qiao, wqiu7, cxliu, alan.yuille\}@jhu.edu \hskip 1em wei.shen@t.shu.edu.cn}
}

\maketitle

\begin{abstract}
Motivated by product detection in supermarkets, this paper studies the problem of object proposal generation in supermarket images and other natural images. We argue that estimation of object scales in images is helpful for generating object proposals, especially for supermarket images where object scales are usually within a small range. Therefore, we propose to estimate object scales of images before generating object proposals. The proposed method for predicting object scales is called ScaleNet. To validate the effectiveness of ScaleNet, we build three supermarket datasets, two of which are real-world datasets used for testing and the other one is a synthetic dataset used for training. In short, we extend the previous state-of-the-art object proposal methods by adding a scale prediction phase. The resulted method outperforms the previous state-of-the-art on the supermarket datasets by a large margin. We also show that the approach works for object proposal on other natural images and it outperforms the previous state-of-the-art object proposal methods on the MS COCO dataset. The supermarket datasets, the virtual supermarkets, and the tools for creating more synthetic datasets will be made public.
\end{abstract}

\section{Introduction}

There is an exciting trend in developing intelligent shopping systems to reduce human intervention and bring convenience to human's life, \emph{e.g.}, \emph{Amazon Go}\footnote{https://www.amazon.com/b?node=16008589011} system, which makes checkout-free shopping experience possible in physical supermarkets.
Another way to enhance the shopping experience in supermarkets is setting customer free from finding and fetching products they want to buy, which drives the demand to develop shopping navigation robots. This kind of robots can also help visually impaired people shop in supermarkets. The vision system of such a robot should have the abilities to address two problems sequentially.
\begin{figure}[!htp]
    \centering
    \includegraphics[width=\linewidth]{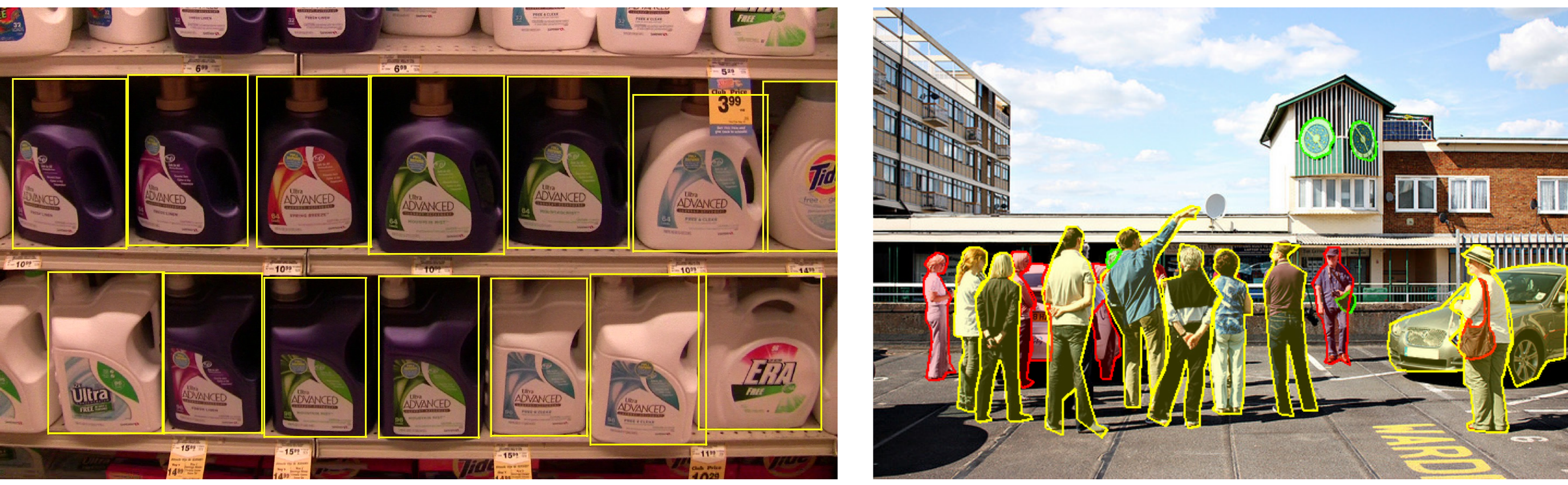}
    \caption{\small Example Object Annotations in the Supermarket Datasets (Left) and the MS COCO Datasets~\cite{DBLP:journals/corr/LinMBHPRDZ14} (Right). Yellow: object scale is between $20\%$ and $30\%$ of the image scale; red: between $10\%$ and $20\%$; green: less than $10\%$. The ratio is calculated as the maximum of the width and the height of the object divided by the maximum of the width and the height of the image.  No other object scales appear in the examples.}
    \label{fig:intropic}
\end{figure}
The first is generating object proposals for products in images captured by the equipped camera (Fig.~\ref{fig:intropic}), and the second is identifying each product proposal. In this paper, we focus on the first problem.

There are many object proposal methods for general natural images \cite{DBLP:conf/nips/PinheiroCD15,pinheiro2016learning,DBLP:journals/ijcv/UijlingsSGS13,DBLP:conf/eccv/ZitnickD14}. However, scenes of supermarkets are usually very crowded, \emph{e.g.}, one image taken in supermarkets could have over $60$ products. More challengingly, products of the same brands and categories are usually placed together, \emph{i.e.}, the appearance similarities between adjacent products are often high, making the boundaries between them hard to detect. Consequently, the current object proposal detection methods, including super-pixel grouping based~\cite{arbelaez2014multiscale,krahenbuhl2014geodesic,DBLP:journals/ijcv/UijlingsSGS13}, edge or gradient computation based~\cite{cheng2014bing,DBLP:conf/eccv/ZitnickD14} and saliency and attention detection based~\cite{borji2015salient,DBLP:conf/iccv/ChangLCL11,DBLP:conf/cvpr/ChangLL11,li2014secrets,liu2011learning}, are less effective and require a large number of proposals to achieve reasonable recall rates.




\begin{figure}[!htp]
    \includegraphics[width=\linewidth]{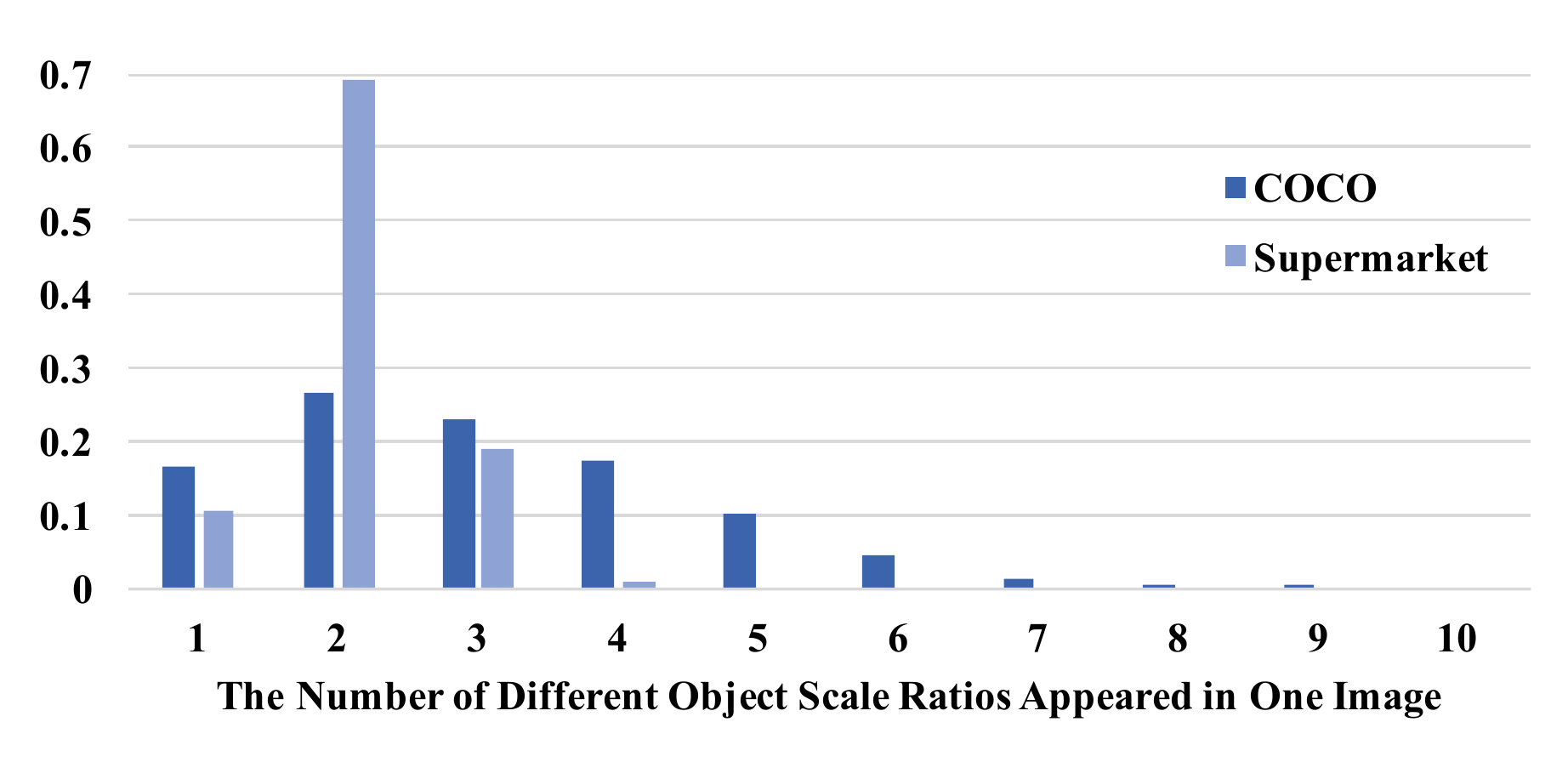}
    \vspace{-0.3in}
    \caption{\small Distributions of the Number of Different Object Scale Ratios of One Image on the MS COCO~\cite{DBLP:journals/corr/LinMBHPRDZ14} Dataset and the Real-World Supermarket Dataset. The ratio of the object size (the maximum of width and height) to the image size (the maximum of width and height) is partitioned evenly to $10$ bins from $0$ to $1$. We count the number of different scale ratios appeared in one image on the datasets. The object scales of supermarket images are sparser than that of images in the MS COCO. Since $97.5\%$ supermarket images have neighboring non-zero bins, the scale distributions are within a small range compared to the entire scale space. Moreover, a reasonable number of images in the MS COCO dataset also have fairly sparse object sizes.}
    \label{fig:scalestat}
\end{figure}

However, we observe that the products in supermarkets typically occur at a limited range of scales in the image.
To demonstrate this, we plot the distribution of the number of object scales in real-world supermarkets (Fig.~\ref{fig:scalestat}).
This suggests a strategy where we estimate object scales and use them to guide proposals rather than exhaustive searching on all scales.
The same strategy of reducing search space of scales is also applicable to other natural images in the MS COCO~\cite{DBLP:journals/corr/LinMBHPRDZ14}, and it becomes very effective especially for those that have sparse object scales (Fig.~\ref{fig:scalestat}), for which an effective scale prediction can reduce the search space and eliminate false positives at improper scales.

\begin{figure*}
    \centering
    \includegraphics[width=\linewidth]{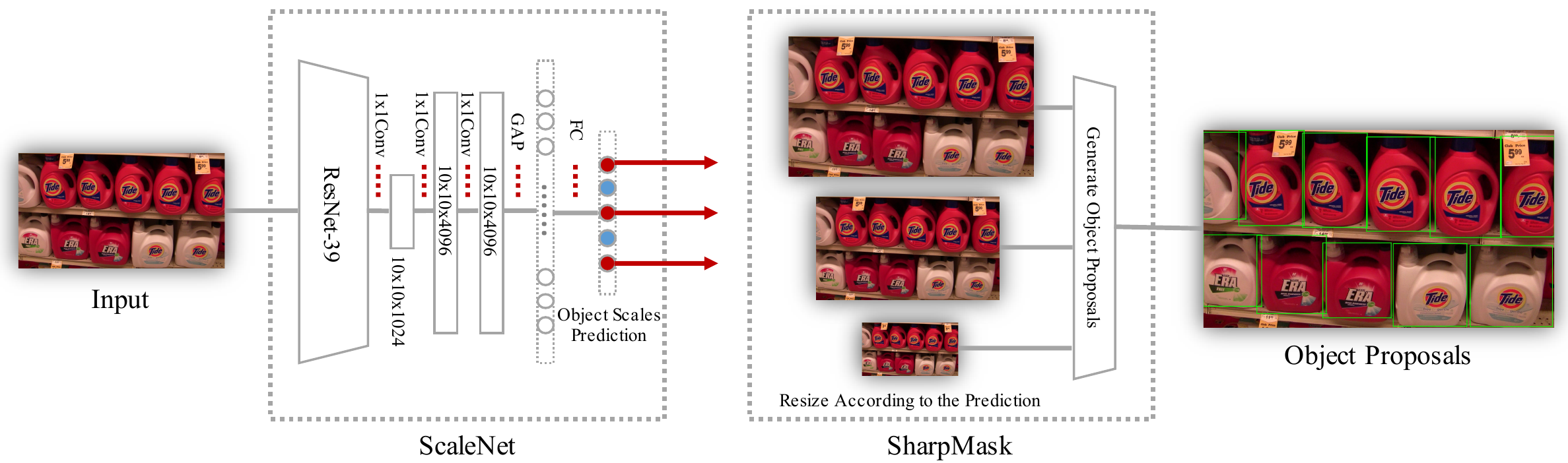}
    \caption{\small The System Overview of the Proposed Object Proposal Framework. The system has two components: ScaleNet proposed in this paper and SharpMask~\cite{pinheiro2016learning}. ScaleNet outputs a predication of the scale distribution of the input image, according to which the input image is resized and fed to SharpMask. SharpMask then generates object proposals at the predicted scales. The image is best viewed in color.}
    \label{fig:overview}
\end{figure*}

More precisely, we propose a scale-aware object proposal detection framework to address the problem (Fig.~\ref{fig:overview}). Our framework consists of two sequential parts. The first is a scale estimation network, called ScaleNet, which predicts the scale distribution of the objects appeared in an image. The second is an object proposal detection network, which performs detection on re-scaled images according to the estimated scales.  For the second part, we use a deep learning based object proposal detection method SharpMask~\cite{pinheiro2016learning}, which predicts objectness confidence scores and object masks at each location of the input image at several pre-defined scales. Since this method can output dense object masks, it fits the supermarket images well.

%
%

We evaluate the proposed framework on general natural images and supermarket images.
To evaluate our framework on natural images, we test it on the MS COCO dataset. For the supermarket images, we collect two real-world supermarket datasets, in which the bounding boxes of products are annotated by humans. The first dataset is called \emph{Real-Far}, which is composed of $4033$ products labeled and has less variation in object scales. The second dataset is called \emph{Real-Near}, which has $3712$ products labeled with more variation in scales. The objective of collecting two datasets is to evaluate and compare the performances in different settings of object scales.

Since human labeling for crowded scenes is very time-consuming and expensive, to generate enough training data, we use a Computer Graphics technique~\cite{DBLP:journals/corr/QiuY16} to generate a synthetic dataset, which includes $154238$ objects labeled for training and $80452$ objects for validation. The synthetic dataset is used for training and validation and the two real-world datasets are used only for testing.

To summarize, the contributions of this paper include
\begin{itemize}[noitemsep,topsep=1.5pt,leftmargin=*]
\item A scale estimation method ScaleNet to predict the object scales of an image.
\item An object proposal framework based on ScaleNet that outperforms the previous state-of-the-arts on the supermarket datasets and MS COCO.
\item Two real-world supermarket datasets and a synthetic dataset, where the model trained only on synthetic dataset transfers well to the real-world datasets. The datasets and the tools will be made public.
\end{itemize}



\section{Related Work}
In this section, we review the related work in the research topics including object proposal methods and virtual environment constructions.

\subsection{Object proposal} The previous work usually falls into two categories: one is bounding box based, and the other is object mask based. Both can generate object proposals in the form of bounding box. In bounding box based methods such as Bing \cite{cheng2014bing} and EdgeBox \cite{DBLP:conf/eccv/ZitnickD14}, local features such as edges and gradients are used for assessing objectness of certain regions.
Following the success of CNNs in image classification \cite{DBLP:conf/cvpr/HeZRS16,krizhevsky2012imagenet,DBLP:journals/corr/SimonyanZ14a},
DeepBox \cite{kuo2015deepbox} re-ranks the object proposals generated by EdgeBox \cite{DBLP:conf/eccv/ZitnickD14}, and DeepProposal \cite{ghodrati2015deepproposal} generates object proposal by an inverse cascade from the final to the initial layer of the CNN.
MultiBox \cite{Erhan_2014_CVPR} and SSD \cite{DBLP:conf/eccv/LiuAESRFB16} compute object regions by bounding box regression based on CNN feature maps directly. In SSD, YOLO \cite{DBLP:conf/cvpr/RedmonDGF16} and RPN \cite{DBLP:conf/nips/RenHGS15}, anchor bounding boxes are used to regress bounding boxes.  Jie {\it et al.} \cite{DBLP:journals/tip/JieLFLTY16} proposed scale-aware pixel-wise proposal framework to handle objects of different scales separately. Although some methods use multi-scales to generate proposals, they do not explicitly estimate the object scales.

Object mask based methods propose object bounding boxes by segmenting the objects of interest from the corresponding background at pixel or region level. This type of methods can detect objects by seed segmentation such as GOP \cite{krahenbuhl2014geodesic} and Learning to Propose Objects \cite{krahenbuhl2015learning}.
They can also group over-segmented regions to propose objects such as Selective Search \cite{DBLP:journals/ijcv/UijlingsSGS13} and MCG \cite{arbelaez2014multiscale}.
More recently, DeepMask \cite{DBLP:conf/nips/PinheiroCD15} assesses objectness and predicts object masks in a sliding window fashion based on CNN features, which achieved the state-of-the-art performance on the PASCAL VOC \cite{everingham2010pascal} and the MS COCO \cite{DBLP:journals/corr/LinMBHPRDZ14} datasets.
SharpMask \cite{pinheiro2016learning} further refines the mask prediction of DeepMask by adding top-down refinement connection. Our method extends the previous state-of-the-art SharpMask by adding object scale prediction and outperforms them on the supermarket dataset and on the MS COCO.

\subsection{Virtual environment construction}
Using synthetic data for Computer Vision research has attracted a lot of attention in recent work. Examples include using synthetic data on semantic segmentation \cite{richter2016playing,ros2016synthia}, optical flow \cite{butler2012naturalistic,DBLP:conf/iccv/DosovitskiyFIHH15},
stereo \cite{mayer2016large,zhang2016unrealstereo}, \emph{etc}. To get virtual environments, the first way is by taking advantages of the existing virtual environments
\cite{DBLP:journals/corr/DosovitskiyK16,johnson2016malmo,DBLP:journals/corr/MahendranBHV16,richter2016playing}.
The second way is to use open source platform such as UnrealCV \cite{DBLP:journals/corr/QiuY16} to construct virtual worlds from scratch. We adopt the second approach and use UnrealCV to build virtual supermarkets.
When constructing virtual environment from scratch, spatial modeling is important for creating realistic environments \cite{Fisher:2012:ESO:2366145.2366154,Yu:2011:MHA:2010324.1964981}. The synthetic dataset introduced in this paper builds the virtual environments from scratch with randomness considered in spatial modeling, material and lighting conditions to create realistic images.

\section{System Overview}
This section presents the system overview of the object proposal framework proposed in this paper, as shown in Fig.~\ref{fig:overview}. The system is composed of two sequential components: the ScaleNet proposed in this paper and SharpMask~\cite{pinheiro2016learning}. The function of ScaleNet is to predict the scales that best describe the statistics of the image so that SharpMask can utilize the predicted scales to find objects better in the image and outputs proposals. ScaleNet looks at the input image only once to predict the distribution of the object scales while SharpMask looks at the input image multiple times at the scales that are predicted by ScaleNet.

The main difference between the proposed framework and SharpMask alone is the way they handle scales. SharpMask exhaustively searches a pre-defined scale set and generates object proposals from that. By contrast, this paper refines the scale set so that SharpMask can take the image at a finer range of scales for object proposal generation.

\section{Scale Distribution Prediction}
This section formulates the problem of scale distribution prediction, presents the architecture of the proposed method ScaleNet, and connects ScaleNet to SharpMask.

\subsection{Problem formalization}
Given an image $I$, we denote the objects of interest in the image $I$ as $O=\{o_1,o_2,...,o_n\}$. Let $m_i$ denote the maximum of the width and the height of the bounding box of object $o_i$, for $i=1,...,n$. Suppose the object $o_i$ can be best detected when the image is resized such that $m_i$ is equal to an ideal size denoted as $D$. This is aiming at work in which there is a set of object sizes that models are trained at \cite{DBLP:journals/corr/ChenYWXY15,DBLP:journals/corr/HuR16,DBLP:conf/nips/PinheiroCD15,pinheiro2016learning,DBLP:journals/corr/XiaWCY15}. Then the scale that image $I$ needs to be resized to favor detecting object $o_i$ is $g_i=D/m_i$. Note that $g_i$ is continuous, and finding scales for every object $o_i$ is inefficient. Therefore, instead of formulating the problem as a regression problem, we discretize the scales into several integer bins and model the problem as a distribution prediction problem.

Suppose for scale distribution we have integer bins $B=\{b_1, b_2, ..., b_l\}$
with discretization precision $\sigma\in\mathbb{Z}^+$, where $b_{i+1}=b_i+1$, $i=1,...,l-1$, and for every possible scale $g_i$ in the dataset $b_1<-\sigma\log_2{g_i}<b_l$.
Then, the ground truth scale distribution $P=\{p_1, p_2, ..., p_l\}$ over the integer bins $B=\{b_1, b_2, ..., b_l\}$ is defined by
\begin{equation}\label{equ:p}
p_i=\frac{\sum_{1\leq j\leq n}{\max{(0, 1 - |b_i+\sigma\log_2{g_j}|)}}}{\sum_{1\leq k\leq l}{\sum_{1\leq j\leq n}{\max{(0, 1 - |b_k+\sigma\log_2{g_j}|)}}}}
\end{equation}

Let $Q=\{q_1,q_2,...,q_l\}$ denote the predicted distribution. We formulate the problem of scale prediction as minimizing Kullback-Leibler divergence (cross entropy) from $Q$ to $P$ defined by
\begin{equation}\label{equ:kl}
D(Q,P)=\sum_{1\leq i\leq l}{p_i\cdot(\log{p_i}-\log{q_i})}
\end{equation}

We now justify Eq.~\ref{equ:p} in details. SharpMask~\cite{pinheiro2016learning} is a scale-sensitive method, which can generate correct object proposals only if the image is properly resized. For each object size, there is a narrow range of image sizes within which the object can be detected. This is where $g_i$ comes from. The rest of Eq. \ref{equ:p} comes naturally.

\subsection{ScaleNet architecture}
To devise a model that outputs $Q$ which minimizes Eq.~\ref{equ:kl}, we propose a deep neural network called ScaleNet. This section presents the architecture of ScaleNet and discusses the motivations behind the design.

The input size of ScaleNet is $192\times192$ with RGB channels. Given input image $I$ of size $w\times h$, we first resize the image to fit the input of ScaleNet $I'$. More specifically, we compute $d=\max(w,h)$, then resize the image such that $d=192$. Next, we copy the resized $I$ to the center of $I'$, and pad $I'$ with a constant value. $I'$ is then fed into ResNet \cite{DBLP:conf/cvpr/HeZRS16} to extract image features. Here, the fully connected layers and the last convolutional stage have been removed from ResNet. After extraction, the features from ResNet go through two $1\times 1$ convolutional stages which serve as local fully connected layers to further process the features separately at each location on the feature map.
ReLU \cite{nair2010rectified} and batch normalization \cite{DBLP:journals/corr/IoffeS15} are used in the two stages to stabilize and speed up training. At the end, a global average pooling layer \cite{DBLP:journals/corr/LinCY13} collects features at each location of the feature map from the two convolutional stages, then outputs scale distribution by a SoftMax operation.

The intuition is to learn the object scales at each location of the image then combine them into one image property. The global average pooling applied at the end of ScaleNet distributes this learning problem to different locations of the image. The distributed tasks can be learned separately by fully connected layers on top of each location of feature map from the last convolutional stage of ResNet. $1\times 1$ convolutional operation then serves as a local fully connected layer to process the features. Similar to the fully connected layers of VGGNet \cite{DBLP:journals/corr/SimonyanZ14a}, we deploy two $4096$ dimension feature extractors. The main difference is that the extracted features in ScaleNet have $4096$ features for each location of feature map instead of the whole image.

\subsection{Connecting ScaleNet to SharpMask}
For an image $I$, ScaleNet is able to predict a scale distribution $Q=\{q_1,...,q_l\}$. This is a probability density function, which we denote as $q(x)$. We assume that the optimal number of scales needed by SharpMask is $h$ (usually $h\sim8$). To exploit $Q$ for SharpMask, the task is to choose a set of scales $S=\{s_1,...,s_h\}$ to resize $I$ as the input of SharpMask. The intuition is to densely sample scales around the scales $b_i$ that have high probability $q_i$. To achieve this, we consider the cumulative distribution function of $q$, \emph{i.e.},
\begin{equation}
    F(s)=\int_{-\infty}^sq(x)~dx
\end{equation}
Then we sample scales in the space of $F(s)$ such that
\begin{equation}
    F(s_i)=\frac{i}{h+1},~\text{for }i=1,...,h
\end{equation}
Before sampling, the distribution $q$ can be smoothed by
\begin{equation}
    q'(x)=\frac{q(x)^\lambda}{\int q(x)^\lambda~dx}
\end{equation}
where $\lambda$ is the smoothing parameter.

\begin{figure*}
\begin{subfigure}{.485\linewidth}
    \centering
    \includegraphics[width=\linewidth]{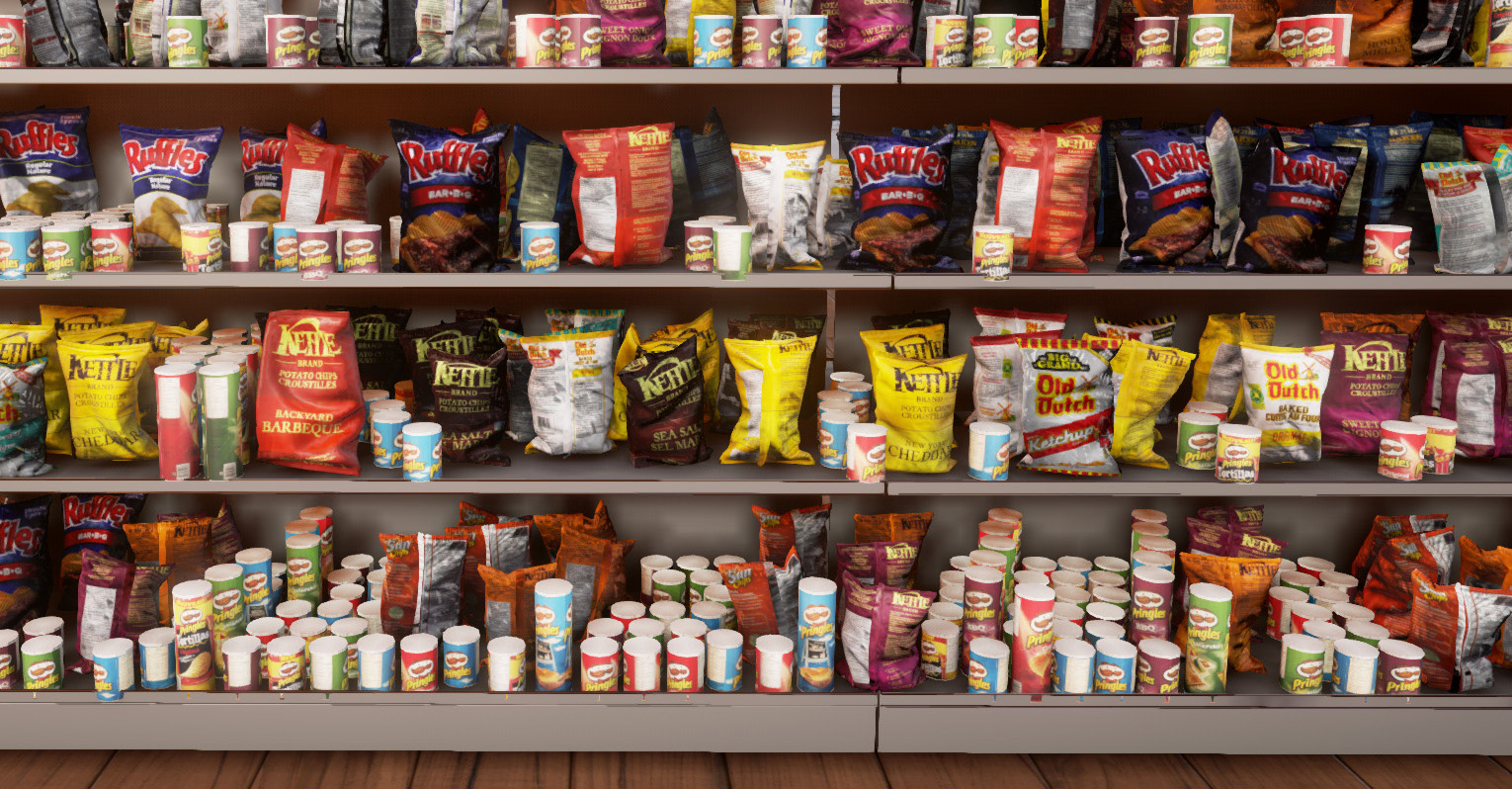}
\end{subfigure}
\hfill
\begin{subfigure}{.485\linewidth}
    \centering
    \includegraphics[width=\linewidth]{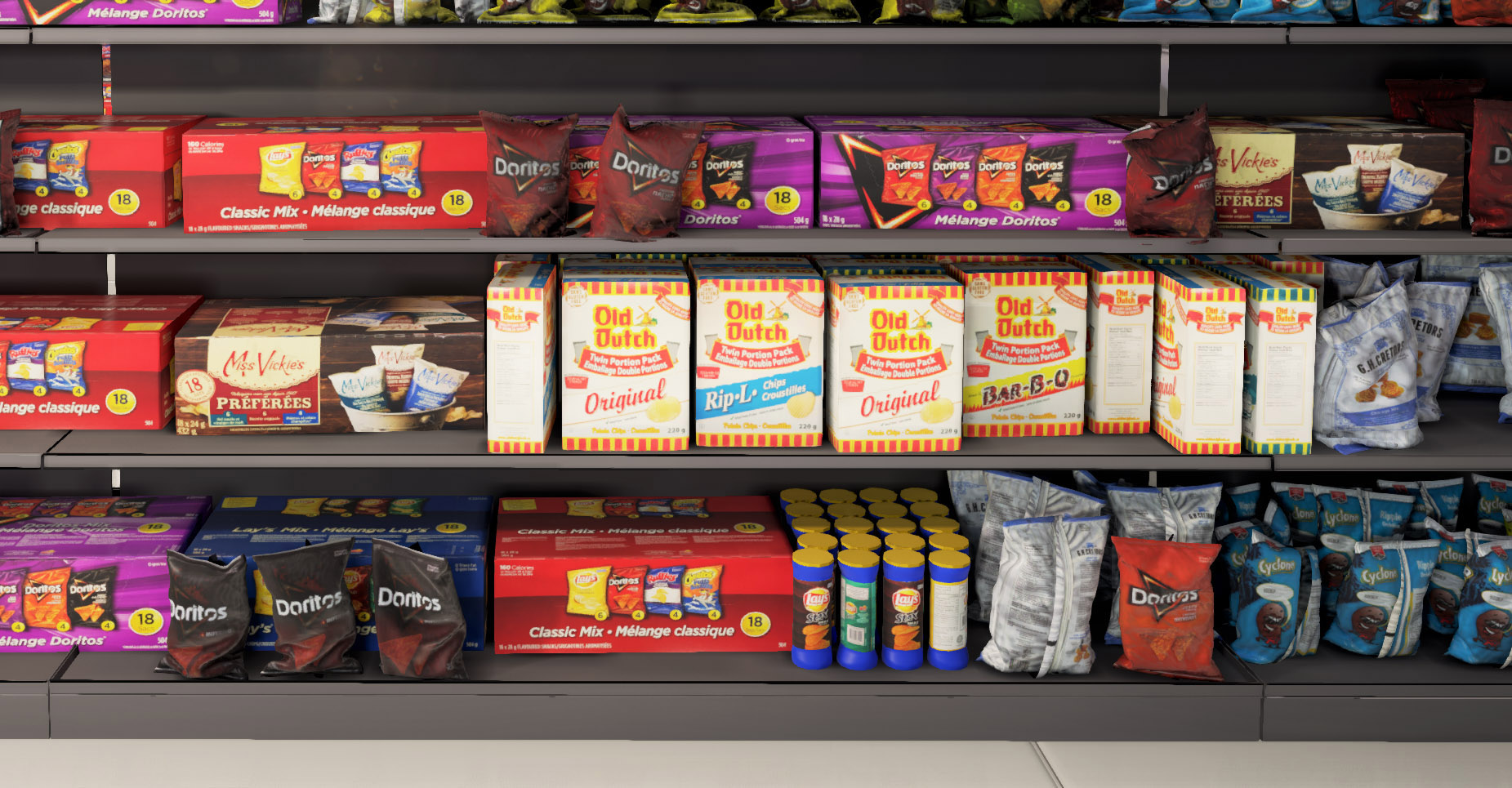}
\end{subfigure}
\caption{\small Comparison of Product Arrangements with Different Proximities. Left: an example of product arrangement result with proximity set to $0$; right: an example of product arrangement result with proximity set to $1$. Setting proximity to a lower value makes the arrangement look more random while setting to a higher value will get a more organized arrangement. The valid range of proximity is within $0$ to $1$.}
\label{fig:randomness}
\end{figure*}

\section{Supermarket Datasets}

\subsection{Real-world datasets}
We aim to study the importance of the scales to the existing object proposal methods; therefore, we prepared two real-world datasets, each of which focuses on one setting of object scales. The first dataset, which we call \emph{Real-Far}, is composed of $4033$ products labeled in bounding boxes. The images in this dataset were taken from a far distance with less variation in scales, thus usually having more objects within one image. On average, one image contains $58$ objects. The second dataset is called \emph{Real-Near}, which contains $3712$ products annotated. For this dataset, we took the images from a near distance and the images have more variation in object scales. The images in \emph{Real-Near} have $27$ products for each on average. Two professional labelers worked on the datasets during collection. In total, we have $7745$ products labeled for testing.

\subsection{Synthetic dataset}
Labeling images in supermarkets can be very time-consuming since there are usually $30$ to $60$ objects in one typical image. Although for SharpMask the number of training examples grows linearly with respect to the number of  the annotated objects, ScaleNet considers one image labeled as one example, thus requiring more data for training; what's more, SharpMask is a mask-based proposal method, which needs objects annotated in object masks, making annotation much harder for humans. Our solution is to build a virtual supermarket to let models learn in this virtual environment. The training and the validation of models are all done in the virtual supermarket. The models are then tested directly on the real-world datasets without fine-tuning. By doing this, we can significantly reduce human labeling, but we need to be very careful when designing the virtual environments so that the models can transfer well to the real-world data from the synthetic data.

\vspace{-0.15in}
\paragraph{Realism}
The first aspect we consider is the realism of the rendered images. Although some work suggested that realism might not be critical for some vision tasks \cite{DBLP:conf/iccv/DosovitskiyFIHH15}, it is a high priority in this paper since we do not fine-tune on the real-world data. The rendering engine we chose is Unreal Engine\footnote{https://www.unrealengine.com/} for its flexibility of object manipulation and high rendering quality. UnrealCV \cite{DBLP:journals/corr/QiuY16} is used to extract the ground truth of object masks. To fully exploit the power of Unreal Engine, all the objects in the virtual supermarket are set to be static and the lighting is baked (\emph{i.e.} pre-computed) before the game is run.

\vspace{-0.15in}
\paragraph{Randomness of placement}
The products in a real supermarket are usually placed according to certain rules. However, since the generalizability must be taken care of when generating a virtual dataset, the randomness of placement is introduced into the rules that guide the construction of the virtual environment.

Similar to some 3D object arrangement methods \cite{Fisher:2012:ESO:2366145.2366154,Yu:2011:MHA:2010324.1964981}, we specify a stochastic grammar of spatial relationship between products and shelves. First, the products are initially located at a position that is not in the rendering range. Next, given a shelf that products can be placed on, the products will be moved to fill the shelf one by one.
Note that similar products are usually placed together in supermarkets. Therefore, before placing the products, for a group of the products, we first find an anchor point on the shelf. Then we specify a parameter, which we call \emph{proximity}, to denote the probability that the next product will be placed near that anchor point or will be placed randomly somewhere on the shelf. Fig. \ref{fig:randomness} demonstrate the placing arrangements with different proximities.

\vspace{-0.15in}
\paragraph{Product overlapping} Product arrangement must prevent overlapping. Motivated by reject sampling, we first randomly create arrangements then reject those that have overlapping products. To efficiently detect overlapping while preserving concave surfaces, convex decomposition is applied to the 3D models before calculating overlapping.

\begin{figure}[!htp]
\centering
\includegraphics[width=0.75\linewidth]{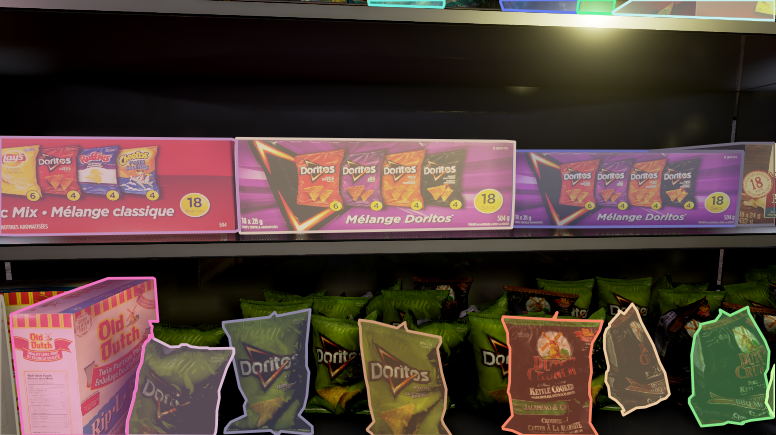}
\caption{\small A Zoom-In Example of the Ground Truth Extracted by UnrealCV \cite{DBLP:journals/corr/QiuY16} with Heavily Occluded Objects Ignored. The virtual dataset is compatible with the MS COCO dataset \cite{DBLP:journals/corr/LinMBHPRDZ14}. The visualization result shown here uses the COCO API. The occlusion threshold is set to $0.9$.}
\label{fig:occ}
\end{figure}

\vspace{-0.15in}
\paragraph{Occlusion} A problem of using synthetic dataset is that all objects will be labeled, including
extremely occluded objects that are usually ignored in building real-world datasets. Our solution to this problem is to calculate the ratio of occlusion for each object, then ignore the objects of occlusion under threshold $\mu$ when extracting the ground truth. To achieve this, we implement a standard rendering pipeline of vertex shader and fragment shader for computing occlusion. To gather data at high speed, we approximate the occlusion calculation by projecting the objects to the surface parallel to the shelf and calculating them only once.

\vspace{-0.15in}
\paragraph{Object scales} The object scales can be controlled by modifying the distance between the camera and the shelf. We set the camera to be at distance $\nu\cdot\text{d}_{\text{max}}$, where $\text{d}_{\text{max}}$ is the distance at which the camera can exactly take in one shelf completely. Then we can modify $\nu$ to generate data with different object scales.

\vspace{-0.15in}
\paragraph{Lighting and material randomness} To augment the virtual dataset, lighting and materials for objects are changed randomly during data gathering.

\vspace{-0.15in}
\paragraph{Summary}
This section presents how the synthetic dataset is constructed with the above aspects taken into account. We develop a plugin for Unreal Engine to construct virtual supermarket stochastically by only one click. We also modify the COCO API to integrate the virtual supermarket dataset into the MS COCO dataset \cite{DBLP:journals/corr/LinMBHPRDZ14}. Fig. \ref{fig:occ} demonstrates the visualization of the mask annotations using the COCO API with the occlusion threshold set to $0.9$.

\section{Implementation Details}
This section presents the implementation details of ScaleNet, the object proposal system, the generation of the virtual supermarket dataset, and the data sampling strategy.

\subsection{Virtual supermarket}
We bought $1438$ 3D models\footnote{https://www.turbosquid.com/} for products and shelves to construct the virtual supermarket. During the data collection, two parameters are manually controlled while others are drawn randomly from a uniform distribution. The two parameters are the occlusion threshold $\mu$ and the distance ratio $\nu$. The range of $\mu$ is \{$0.9$, $0.8$, $0.7$, $0.6$, $0.5$\}, and the range of $\nu$ is \{$1$, $1/1.5$, $1/2$, $1/2.5$, $1/3$\}. Combining different $\mu$ and different $\nu$ results in $25$ configurations, for each we use different product arrangements, and random lighting/material settings at each frame to generate $200$ images.
The above process generates $5000$ synthetic images and $234690$ objects labeled in total. We denote this virtual dataset as dataset {\it V}. We split dataset {\it V} into {\it Vtrain} and {\it Vval} for training and validation, respectively.  The dataset {\it Vtrain} has $3307$ images and $154238$ objects while the dataset {\it Vval} has $1693$ images and $80452$ objects.

\subsection{ScaleNet}
We use Torch7 to build and test ScaleNet. Before training ScaleNet, the ResNet component is pre-trained on ImageNet \cite{ILSVRC15}. The discretization precision $\sigma$ is set to $1$, while the discrete scale bins are set to $B=\{-32,-31,...,0,...,31,32\}$. To accommodate the parameters used in SharpMask \cite{pinheiro2016learning}, $D$ is set to $640/7$.

During training, we resize the image to fit the input of ScaleNet, and calculate the scale distribution $P$ as the ground truth. The mean pixel calculated on ImageNet is subtracted from input image before feeding into ScaleNet. All layers are trained, including the ResNet component. We train two ScaleNet models for the supermarket datasets and the MS COCO \cite{DBLP:journals/corr/LinMBHPRDZ14} dataset, individually. We use the corresponding models when evaluating the performances on different datasets. The training dataset for ScaleNet for supermarket datasets is {\it COCOtrain} + {\it Vtrain} while the validation dataset is {\it COCOval} + {\it Vval}. For the MS COCO, the datasets used for training and validation include only the MS COCO itself. Here, {\it COCOtrain} and {\it COCOval} are the training and the validation set of the MS COCO, respectively.
To connect ScaleNet to SharpMask, $h$ is set to $6$ for the supermarket datasets, and $10$ for the MS COCO. The smoothing factor $\lambda$ is set to $0.9$ for the supermarket datasets, and $0.25$ for the MS COCO.

\subsection{Data sampling}
In the original data sampling strategy adopted in both DeepMask and SharpMask, each image has the same probability for objectness score training and each category has the same probability for object mask training. Instead, we propose to train both the objectness score and object mask so that each annotation has the same probability of being sampled. Following this strategy, the performance can be slightly improved. We denote SharpMask trained in this way as SharpMask-Ours.

\begin{figure*}
    \centering
    \includegraphics[width=\linewidth]{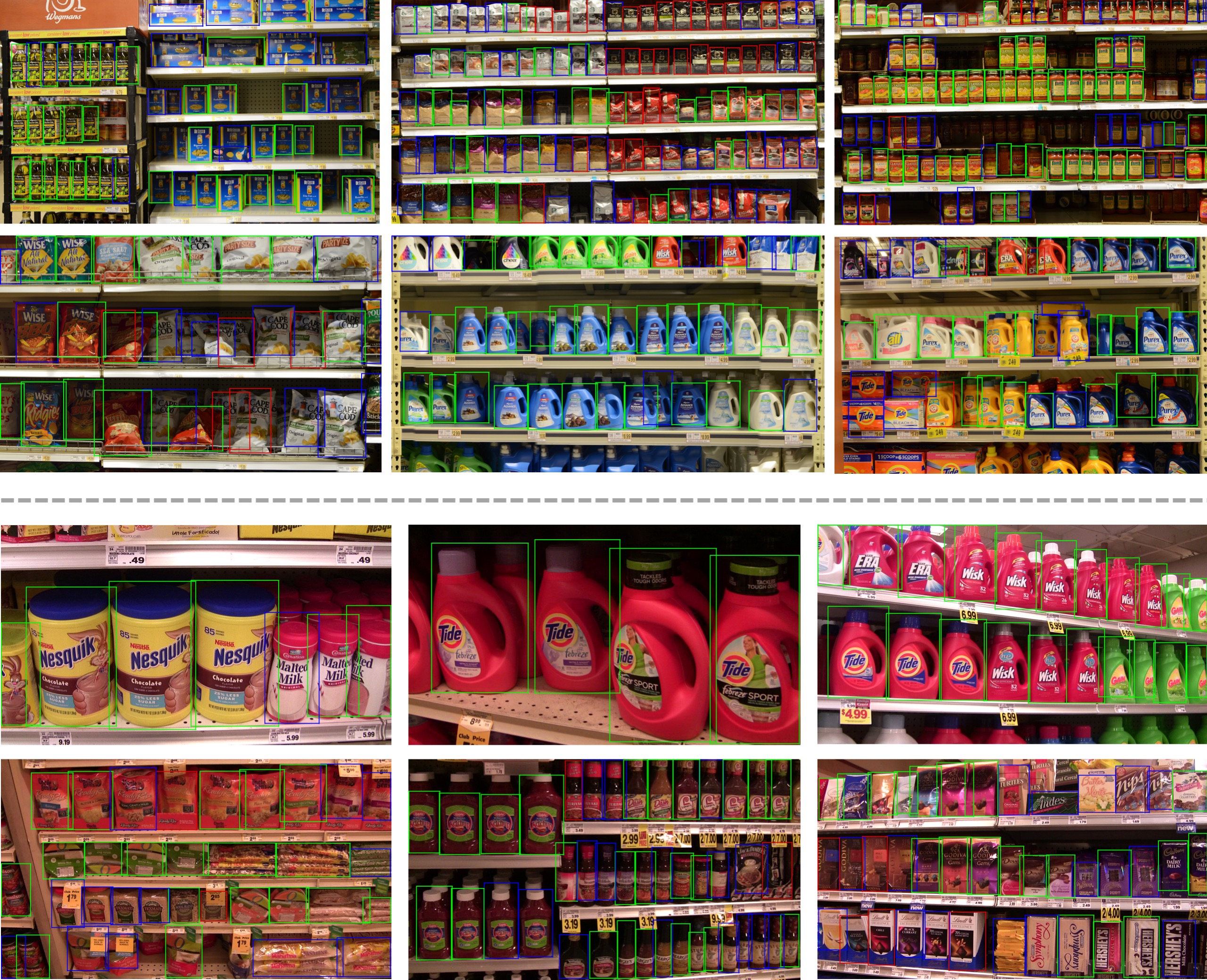}
    \caption{\small Proposals Generated by Our Method ScaleNet+SharpMask-ft with Highest IoU to the Ground Truth on the Selected Real-World Supermarket Images. Top images are selected from dataset \emph{Real-Far} while bottom images are selected from dataset \emph{Real-Near}. Green bounding boxes are from top $100$ proposals. Blue bounding boxes are from proposals ranked between $101$ and $1000$. Red bounding boxes are ground truth of objects not found by our method within $1000$ proposals. The IoU threshold is set to $0.7$.}
    \label{fig:demo}
\end{figure*}

\begin{figure*}
\begin{subfigure}{.333\linewidth}
    \centering
    \includegraphics[width=\linewidth]{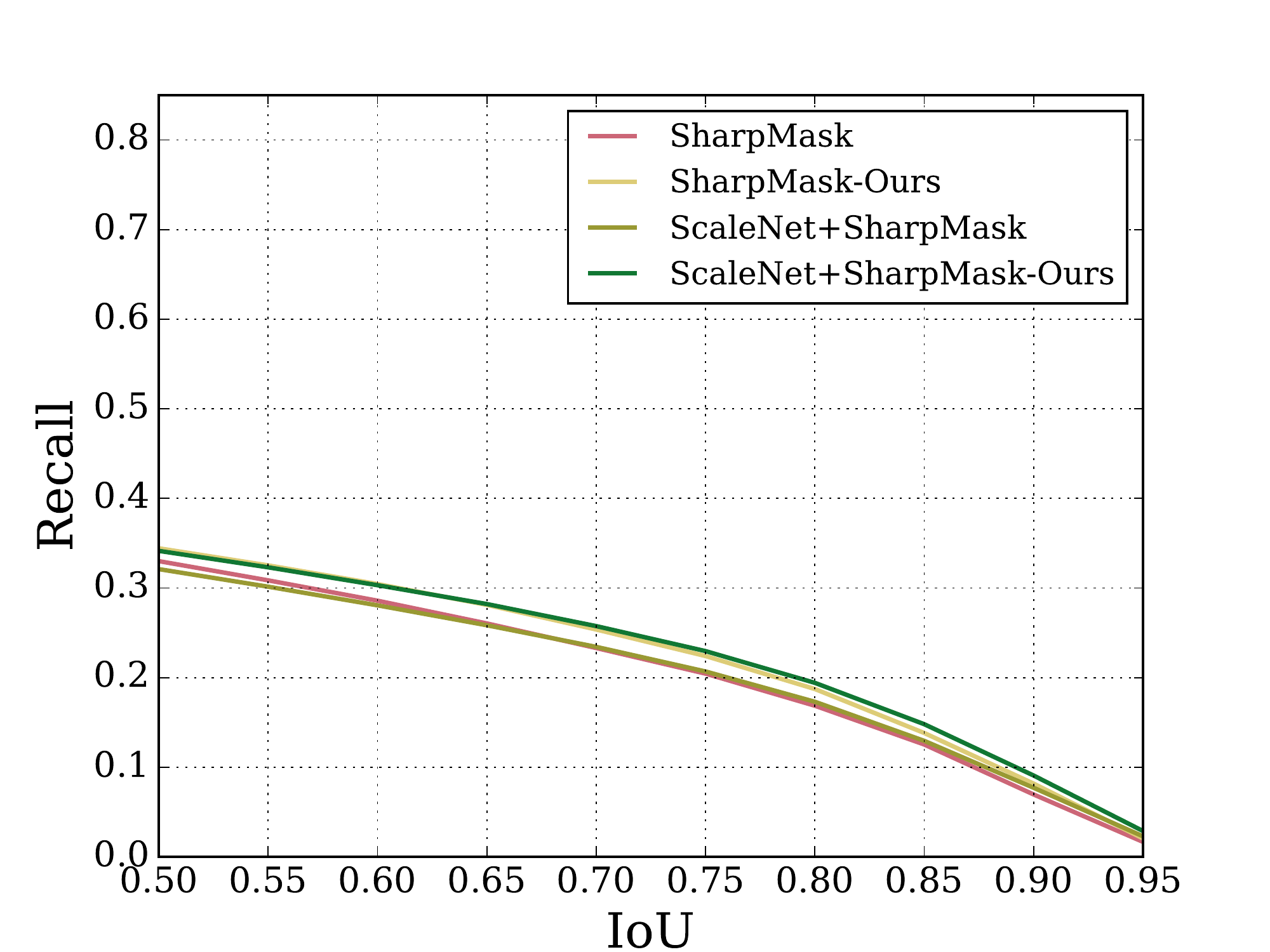}
    \caption{Recall @10 Proposals}
\end{subfigure}
\begin{subfigure}{.333\linewidth}
    \centering
    \includegraphics[width=\linewidth]{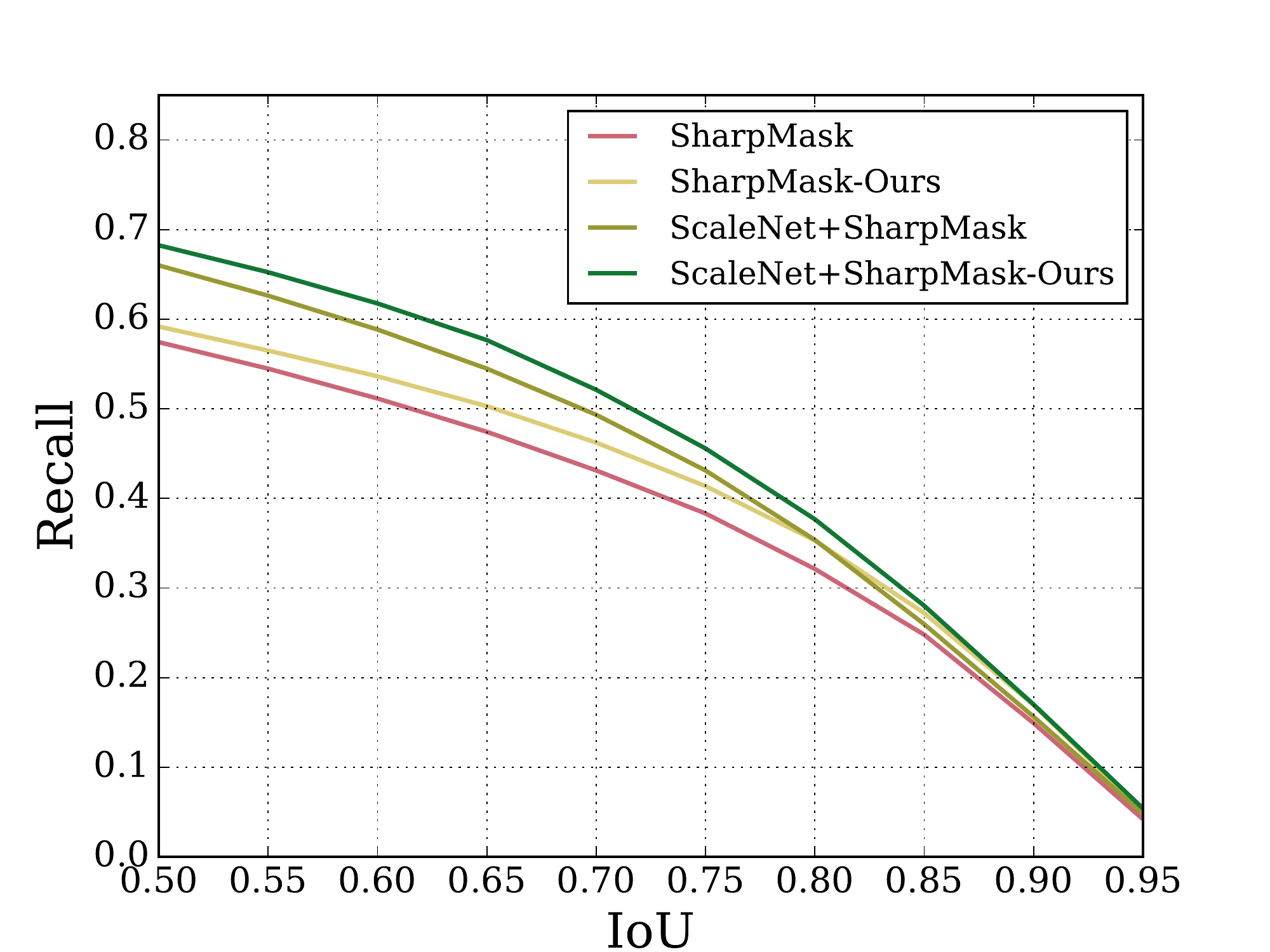}
    \caption{Recall @100 Proposals}
\end{subfigure}
\begin{subfigure}{.333\linewidth}
    \centering
    \includegraphics[width=\linewidth]{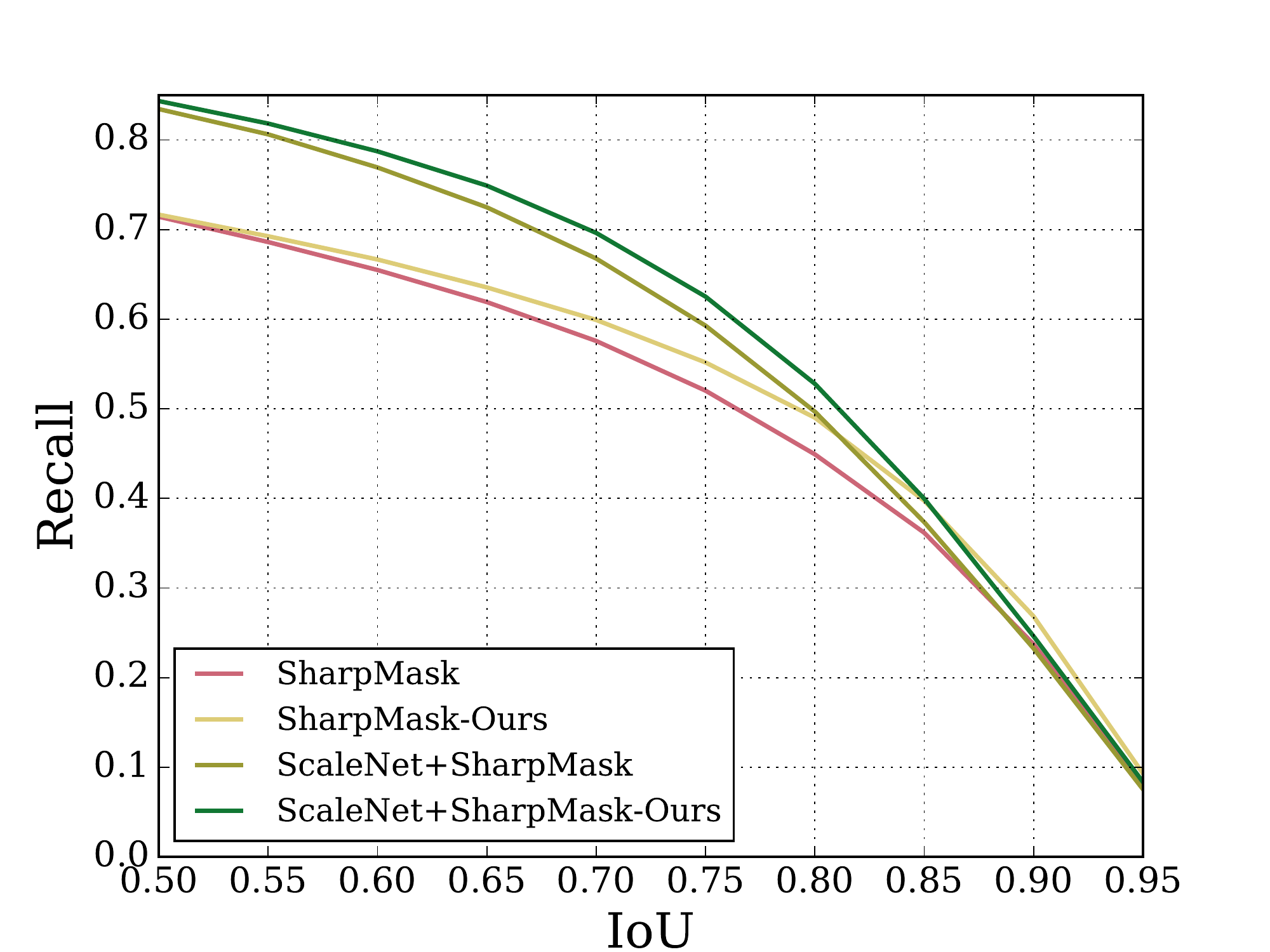}
    \caption{Recall @1000 Proposals}
\end{subfigure}
\caption{Recall versus IoU Threshold for Different Number of Bounding Box Proposals on the MS COCO Dataset.}
\label{fig:curve}
\end{figure*}

\begin{figure*}
\begin{subfigure}{.333\linewidth}
    \centering
    \includegraphics[width=\linewidth]{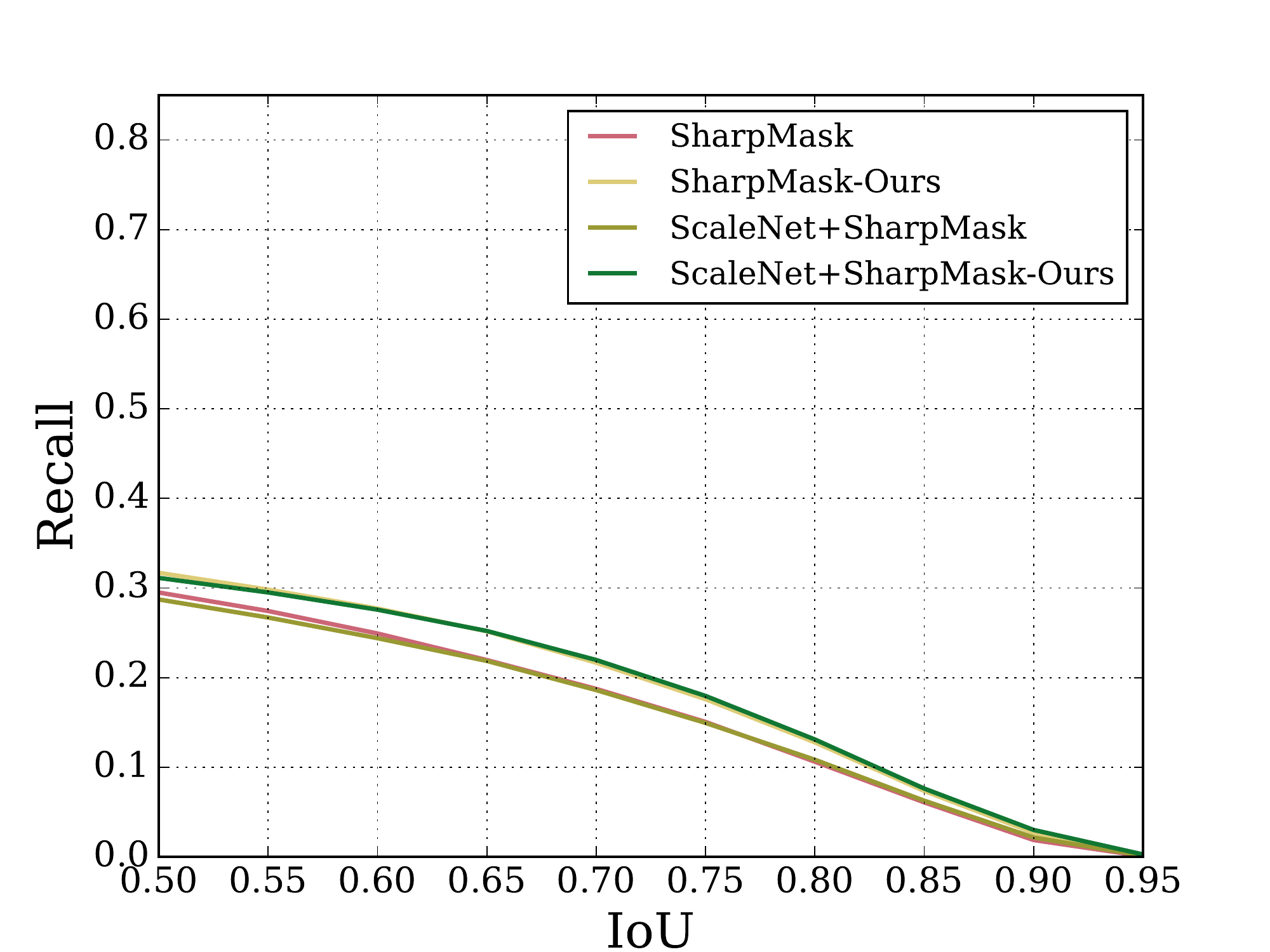}
    \caption{Recall @10 Proposals}
\end{subfigure}
\begin{subfigure}{.333\linewidth}
    \centering
    \includegraphics[width=\linewidth]{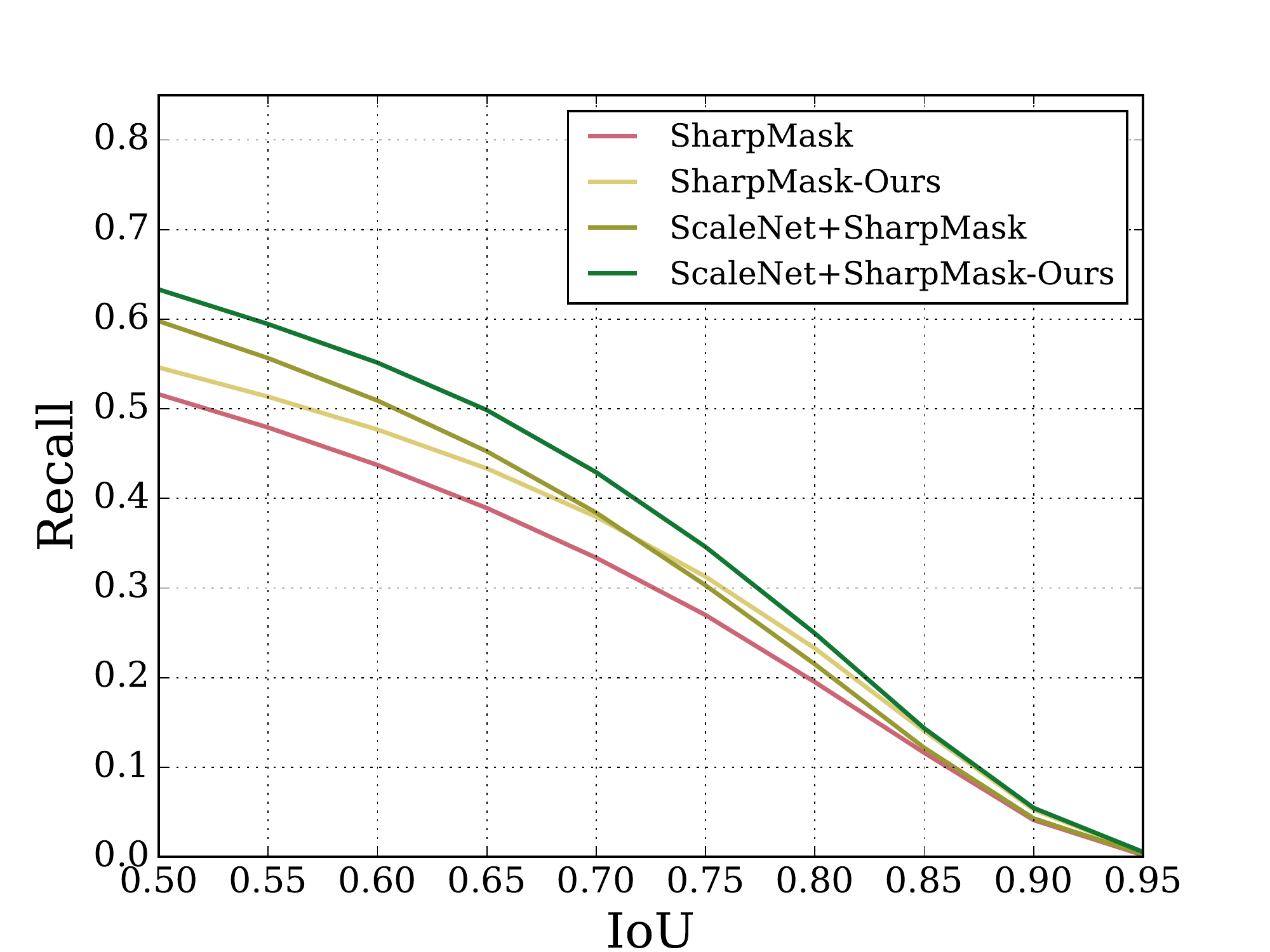}
    \caption{Recall @100 Proposals}
\end{subfigure}
\begin{subfigure}{.333\linewidth}
    \centering
    \includegraphics[width=\linewidth]{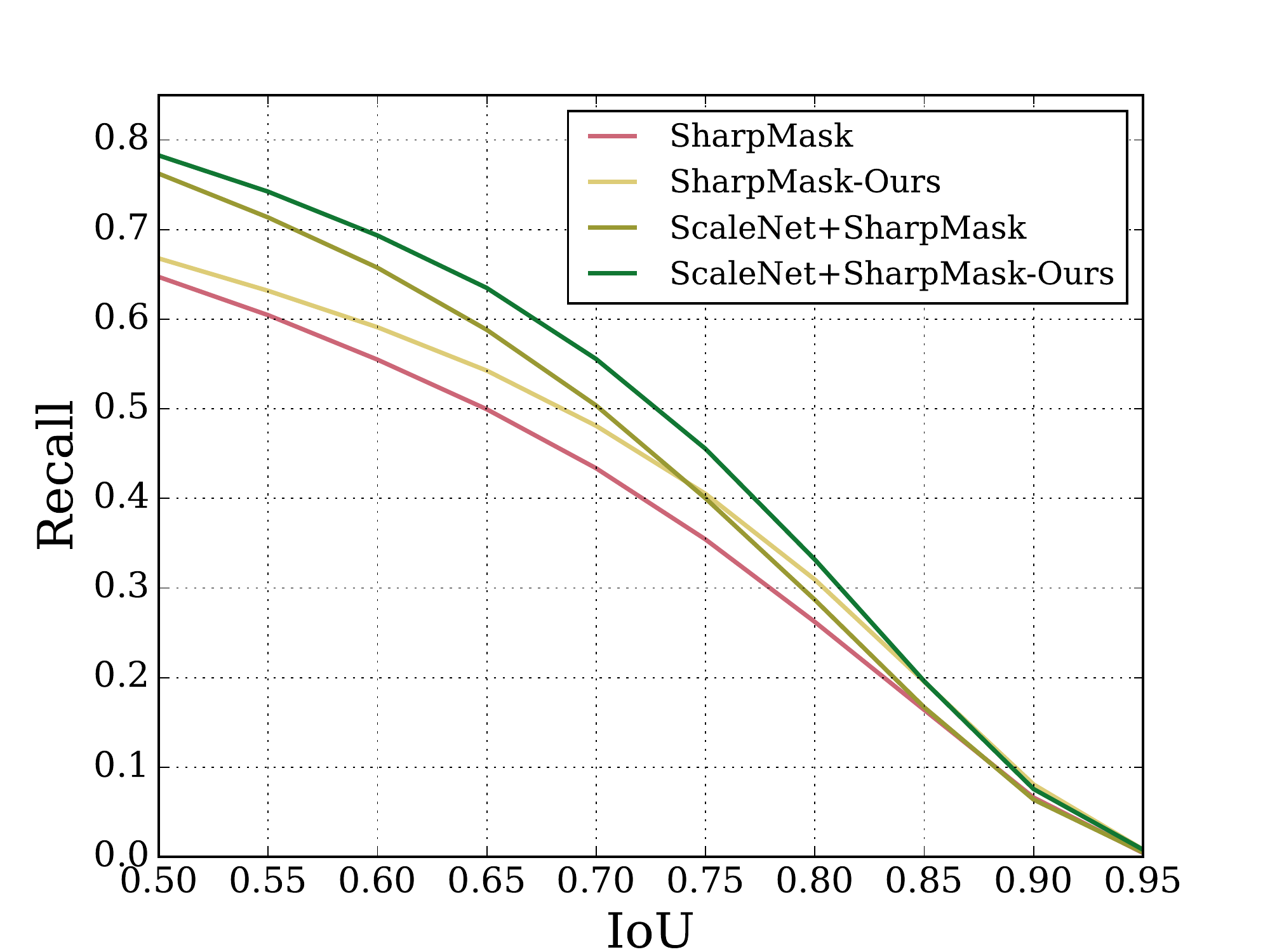}
    \caption{Recall @1000 Proposals}
\end{subfigure}
\caption{Recall versus IoU Threshold for Different Numbers of Segmentation Proposals on the MS COCO Dataset.}
\label{fig:curvemask}
\end{figure*}

\section{Experimental Results}

\begin{table}
\small
\centering
\begin{tabular}{lcc}
\toprule
Methods & \it Real-Far  & \it Real-Near \\
\midrule
EdgeBox@100 \cite{DBLP:conf/eccv/ZitnickD14} &
0.006 &
0.015 \\
Selective Search@100 \cite{DBLP:journals/ijcv/UijlingsSGS13} &
0.019 &
0.043 \\
DeepMask@100 \cite{DBLP:conf/nips/PinheiroCD15} &
0.183 &
0.198 \\
SharpMask@100 \cite{pinheiro2016learning} &
0.191 &
0.205 \\
\hdashline
DeepMask-ft@100 &
0.209 &
0.231 \\
SharpMask-ft@100 &
0.224 &
0.249 \\
ScaleNet+DeepMask@100 &
0.256 &
0.342 \\
ScaleNet+DeepMask-ft@100 &
0.278 &
0.373 \\
ScaleNet+SharpMask@100 &
0.269 &
0.361 \\
ScaleNet+SharpMask-ft@100 &
\bf 0.298 &
\bf 0.396 \\
\midrule
EdgeBox@1000 &
0.203 &
0.324 \\
Selective Search@1000 &
0.225 &
0.328 \\
DeepMask@1000 &
0.472 &
0.488 \\
SharpMask@1000 &
0.499 &
0.518 \\
\hdashline
DeepMask-ft@1000 &
0.497 &
0.533 \\
SharpMask-ft@1000 &
0.526 &
0.567 \\
ScaleNet+DeepMask@1000 &
0.542 &
0.593 \\
ScaleNet+DeepMask-ft@1000 &
0.561 &
0.621 \\
ScaleNet+SharpMask@1000 &
0.570 &
0.625 \\
ScaleNet+SharpMask-ft@1000 &
\bf 0.589 &
\bf 0.651 \\
\bottomrule
\end{tabular}
\caption{\small The Comparison of the Average Recalls \cite{DBLP:journals/corr/HosangBDS15} of Object Proposal Methods Tested on the Real-World Supermarket Datasets {\it Real-Far} and {\it Real-Near}. The method name indicates what method is used and how many proposals are considered in computing recall rates, \emph{e.g.}, EdgeBox@100 means EdgeBox with the number of object proposals limited to $100$. Methods that have suffix \emph{-ft} are trained on the MS COCO and the synthetic supermarket dataset.}
\label{tab:realworld}
\end{table}

\subsection{Object proposal on supermarket datasets}
We first present the performance of our model on the supermarket datasets while only trained on the combination of the MS COCO training dataset and the virtual supermarket training dataset. We evaluated the methods on the dataset \emph{Real-Near} and \emph{Real-Far}. Qualitative results of our method are shown in Fig. \ref{fig:demo}.

\vspace{-0.15in}
\paragraph{Metrics}
The metric used to evaluate the performance of the object proposal methods is the Average Recalls (AR) \cite{DBLP:journals/corr/HosangBDS15} over $10$ intersection over union thresholds from $0.5$ to $0.95$ with $0.05$ as step length.

\vspace{-0.15in}
\paragraph{Methods}
We compare the performance of the proposed method with the top methods of proposing bounding boxes for objects: DeepMask \cite{DBLP:conf/nips/PinheiroCD15}, SharpMask \cite{pinheiro2016learning}, Selective Search \cite{DBLP:journals/ijcv/UijlingsSGS13}, and EdgeBox \cite{DBLP:conf/eccv/ZitnickD14}.

\vspace{-0.15in}
\paragraph{Model transferability}
Table \ref{tab:realworld} demonstrates the improvements of performances of the model trained using virtual supermarket dataset.  Methods that have suffix \emph{-ft} are
trained on the MS COCO and the synthetic supermarket dataset. It's worth noting that the models trained solely on the combination of the general purpose dataset and the task specific synthetic dataset exhibit consistent improvements on the task specific real-world datasets even none of them has a look at the real-world data.

\vspace{-0.15in}
\paragraph{Scales}
Table \ref{tab:realworld} compares the different object proposal methods on the two real-world dataset {\it Real-Near} and {\it Real-Far}.
Without the help of ScaleNet to narrow down the search space of scales, DeepMask and SharpMask actually have similar performances on them. Instead, our proposed method exhibit stronger improvements on \emph{Real-Near} in which the image has fewer objects, thanks to the accurate prediction by ScaleNet of the scales to resize images.

In short, Table \ref{tab:realworld} demonstrates the significant performance improvements by using our proposed framework.

\subsection{Object proposal on the MS COCO dataset}
Next, we evaluate our method on the MS COCO dataset. Following the evaluations done in DeepMask \cite{DBLP:conf/nips/PinheiroCD15} and SharpMask \cite{pinheiro2016learning}, the recall rates are evaluated on the first 5000 images on the validation set.

\vspace{-0.15in}
\paragraph{Methods}
We compare the performance of the proposed method with the state-of-the-art methods of proposing bounding boxes for objects: DeepMask-VGG \cite{DBLP:conf/nips/PinheiroCD15}, DeepMaskZoom-VGG \cite{DBLP:conf/nips/PinheiroCD15}, DeepMask-Res39~\cite{pinheiro2016learning}, SharpMask \cite{pinheiro2016learning}, SharpMaskZoom \cite{pinheiro2016learning}. For segmentation proposals, we also show the comparison with Instance-Sensitive FCN \cite{DBLP:conf/eccv/DaiHLR016}.

\vspace{-0.15in}
\paragraph{Metrics}
We adopt the same metrics used for evaluating performances on the supermarket datasets. The performances are evaluated when the number of proposals is limited to $10$, $100$ and $1000$.

\begin{table}
\small
\centering
\begin{tabular}{lccc}
    \toprule
    Methods & AR@10 & AR@100 & AR@1k \\
    \midrule
    DeepMask-VGG \cite{DBLP:conf/nips/PinheiroCD15} & 0.153 & 0.313 & 0.446 \\
    DeepMaskZoom-VGG \cite{DBLP:conf/nips/PinheiroCD15} & 0.150 & 0.326 & 0.482 \\
    DeepMask-Res39 \cite{pinheiro2016learning} & 0.180 & 0.348 & 0.470 \\
    SharpMask \cite{pinheiro2016learning} & 0.197 & 0.364 & 0.482 \\
    SharpMaskZoom \cite{pinheiro2016learning} & 0.201 & 0.394 & 0.528 \\
    \hdashline
    SharpMask-Ours & 0.216 & 0.392 & 0.510 \\
    ScaleNet+SharpMask & 0.201 & 0.416 & 0.557 \\
    ScaleNet+SharpMask-Ours & \bf 0.220 & \bf 0.439 & \bf 0.578 \\
    \bottomrule
\end{tabular}
\caption{Comparison of Our Framework to DeepMask \cite{DBLP:conf/nips/PinheiroCD15} and SharpMask  \cite{pinheiro2016learning} on Bounding Box Object Proposals on the MS COCO validation dataset \cite{DBLP:journals/corr/LinMBHPRDZ14}.}
\label{tab:coco}
\end{table}

\begin{table}
\small
\centering
\begin{tabular}{lccc}
    \toprule
    Methods & AR@10 & AR@100 & AR@1k \\
    \midrule
    DeepMask-VGG \cite{DBLP:conf/nips/PinheiroCD15} & 0.126 & 0.245 & 0.331 \\
    DeepMaskZoom-VGG \cite{DBLP:conf/nips/PinheiroCD15} & 0.127 & 0.261 & 0.366 \\
    DeepMask-Res39 \cite{pinheiro2016learning} & 0.144 & 0.258 & 0.331 \\
    SharpMask \cite{pinheiro2016learning} & 0.156 & 0.276 & 0.355 \\
    SharpMaskZoom \cite{pinheiro2016learning} & 0.161 & 0.303 & 0.392 \\
    Instance-Sensitive FCN \cite{DBLP:conf/eccv/DaiHLR016} & 0.166 & 0.317 & 0.392 \\
    \hdashline
    SharpMask-Ours & 0.177 & 0.309 & 0.391 \\
    ScaleNet+SharpMask & 0.155 & 0.319 & 0.415 \\
    ScaleNet+SharpMask-Ours & \bf 0.177 & \bf 0.351 & \bf 0.448 \\
    \bottomrule
\end{tabular}
\caption{Comparison of Our Framework with DeepMask \cite{DBLP:conf/nips/PinheiroCD15} and SharpMask  \cite{pinheiro2016learning} on Segmentation Proposals on the MS COCO dataset \cite{DBLP:journals/corr/LinMBHPRDZ14}.}
\label{tab:cocomask}
\end{table}

\vspace{-0.15in}
\paragraph{Results}
Table \ref{tab:coco} and Table \ref{tab:cocomask} summarize the performance comparisons on the MS COCO dataset. Since the object scales in these natural images are not always sparse, we do not expect significant improvements as shown in the supermarket datasets. However, consistent improvements can be observed at all number of proposals.

Fig.~\ref{fig:curve} and Fig.~\ref{fig:curvemask} show the additional performance plots comparing our methods with the previous state-of-the-art. Our framework improves the recall rates significantly at $1000$ proposals, \emph{e.g.}, for bounding box object proposal, the recall rate increases from $0.714$ to $0.843$ when IoU threshold is set to $0.5$, and from $0.575$ to $0.696$ at $0.7$ IoU threshold.  We also observe strong performance increases at $100$ proposals: the recall rate at $0.5$ IoU threshold increases from $0.574$ to $0.682$, and from $0.431$ to $0.521$ at $0.7$ IoU threshold.

\section{Conclusion and Future Work}
In this paper, we study the problem of object proposal generation in supermarket images and other natural images. We introduce three supermarket datasets -- two real-world datasets and one synthetic dataset. We present an innovative object proposal framework, in which the object scales are first predicted by the proposed scale prediction method ScaleNet. The experimental results demonstrate that the model trained solely on the combination of the MS COCO dataset and the synthetic supermarket dataset transfers well to the two real-world supermarket datasets. The proposed scale-aware object proposal method is evaluated on the real-world supermarket datasets and the MS COCO dataset. Our proposed method outperforms the previous state-of-the-art by a large margin on these datasets for the task of object detection in the form of bounding box.

In the future work, since the strategy of reducing search space of object scales is also applicable to other object proposal methods, it is of interest to study how to connect ScaleNet with other methods. Moreover, analyzing what features ScaleNet has learned is also helpful for understanding the structures of natural images.

{\small
\bibliographystyle{ieee}
\bibliography{egbib}
}

\end{document}